\documentclass[10pt,journal,compsoc]{IEEEtran}
\usepackage[left=1.5cm,right=1.5cm,top=2cm,bottom=2cm]{geometry}
\usepackage{ragged2e}
\justifying

\ifCLASSOPTIONcompsoc
   \usepackage[nocompress]{cite}
\else
    \usepackage{cite}
\fi

\ifCLASSINFOpdf
 
\else
 
\fi
\usepackage{multirow}
\usepackage{graphicx}
\usepackage{xcolor}
\usepackage{tabularx}

\usepackage{authblk}

\begin{document}

\title{Types of Approaches, Applications and Challenges in the Development of Sentiment Analysis Systems}

% \author{Lulit Asfaw}
% \author{Mikael Clemmons}
% \author{Cody Hayes}
% \author{Elise Letnaunchyn}
% \affil{College of Computing and Software Engineering,\\ Kennesaw State University, GA, USA
%  \\{lasfaw, mclemmo2, chayes86, eletnaun}@students.kennesaw.edu
% }

\author{\IEEEauthorblockN{Kazem Taghandiki\IEEEauthorrefmark{1},
Elnaz Rezaei Ehsan\IEEEauthorrefmark{2}}
	\IEEEauthorblockA{
	 %	\IEEEauthorrefmark{1}Computer Science and Engineering Department, Indian Institute of Technology Guwahati, Guwahati, India, email: zolfaghari.b1975@gmail.com \\
	 \\
		\IEEEauthorrefmark{1}Department of Computer Engineering, \\Technical and Vocational University (TVU), Tehran, Iran
 \\ktaghandiki@tvu.ac.ir
 \\
					\IEEEauthorrefmark{2} Master's degree, industrial engineering,\\ System management and productivity, Iran University of Science and Technology\\  elnazrezaeie110@gmail.com  \\
							}
     
							}

%\author{Michael~Shell,~\IEEEmembership{Member,~IEEE,}
        %John~Doe,~\IEEEmembership{Fellow,~OSA,}
        %and~Jane~Doe,~\IEEEmembership{Life~Fellow,~IEEE}% <-this % stops a space
%\IEEEcompsocitemizethanks{\IEEEcompsocthanksitem M. Shell was with the Department
%of Electrical and Computer Engineering, Georgia Institute of Technology, Atlanta,
%GA, 30332.\protect\\

%E-mail: see http://www.michaelshell.org/contact.html
%\IEEEcompsocthanksitem J. Doe and J. Doe are with Anonymous University.}% <-this % stops an unwanted space

%{Shell \MakeLowercase{\textit{et al.}}: Bare Demo of IEEEtran.cls for Computer Society Journals}

\IEEEtitleabstractindextext{%
\begin{abstract}
\textcolor{black}{Today, the web has become a mandatory platform to express users' opinions, emotions and feelings about various events. Every person using his smartphone can give his opinion about the purchase of a product, the occurrence of an accident, the occurrence of a new disease, etc. in blogs and social networks such as (Twitter, WhatsApp, Telegram and Instagram) register. Therefore, millions of comments are recorded daily and it creates a huge volume of unstructured text data that can extract useful knowledge from this type of data by using natural language processing methods. Sentiment analysis is one of the important applications of natural language processing and machine learning, which allows us to analyze the sentiments of comments and other textual information recorded by web users. Therefore, the analysis of sentiments, approaches and challenges in this field will be explained in the following. }
\end{abstract}

}

\maketitle

\IEEEdisplaynontitleabstractindextext

\IEEEpeerreviewmaketitle

\section{Introduction}
Today, web users freely register their feedback in blogs, forums and various social networks (such as Twitter, WhatsApp, Telegram, Instagram) \cite{w1}. This type of way of recording opinions has caused the formation of a large volume of unstructured textual data. Therefore, researchers have always sought to provide solutions to extract and discover useful and usable knowledge from this type of unstructured data using natural language processing and machine learning methods. Sentiment analysis is one of the applications of natural language processing that can show the opinions and feedback of web users about an event in the form of (positive feeling, negative feeling and neutral feeling) \cite{1}. Sentiment analysis is like a classification problem that can be processed using different supervised learning algorithms, although in recent years, researchers have used unsupervised learning techniques to solve these types of problems \cite{2,a1}.\\
Sentiment analysis has 3 different processing levels as shown in Figure 1.
   \begin{figure}
    \centering
    \includegraphics[width=7cm,height=7cm]{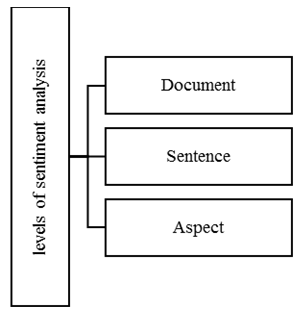}
    \caption{Different levels of sentiment analysis}
    \label{fig:life}
\end{figure}\\
\textbf{A: sentiment analysis at the document level}\\
At this level, the researcher analyzes the feelings of the text in the entire document to map a document to a feeling (positive, negative and neutral) using natural language processing and machine learning algorithms.\\
\textbf{B: Sentiment analysis at the sentence level}\\
At this level, the researcher analyzes the feelings of each sentence in the document to map each sentence to a sense (positive, negative and neutral) using natural language processing and machine learning algorithms.\\
\textbf{C: Analyzing emotions at the aspect or feature level}\\
At this level, the researcher analyzes the feelings of different aspects of a text or sentence (Figure 2). For example: "The story of the movie was not very good, but the music was very good".
   \begin{figure*}
    \centering
    \includegraphics{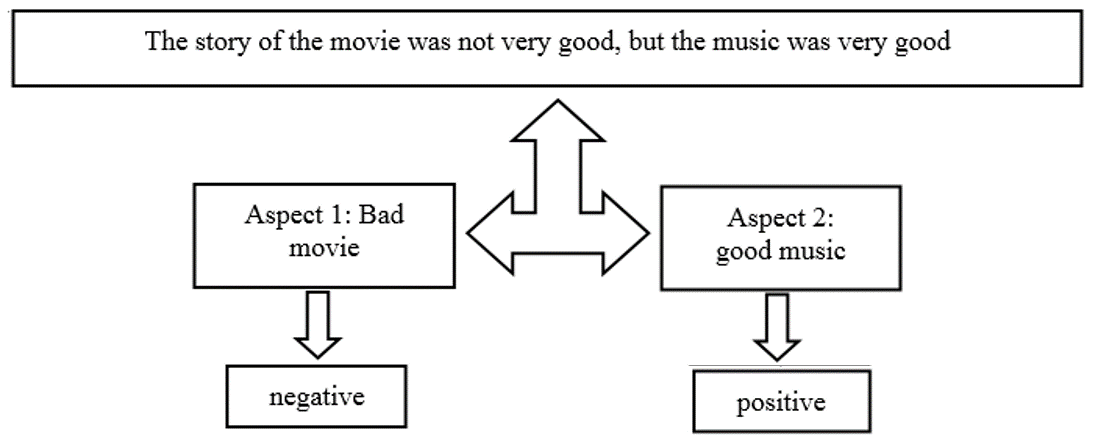}
    \caption{Sentiment analysis at the aspect or feature level}
    \label{fig:life}
\end{figure*}
Figure 3 shows the types of goals in a sentiment analysis system.
 \begin{figure}
    \centering
    \includegraphics[width=9cm]{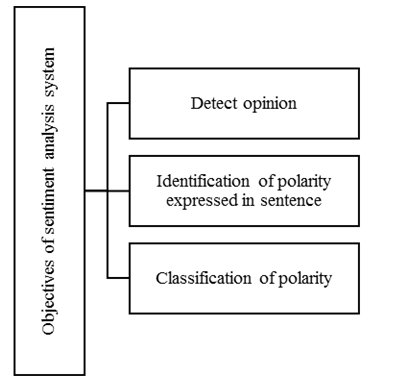}
    \caption{Objectives of sentiment analysis system}
    \label{fig:life}
\end{figure} 

\section{The process of sentiment analysis }\label{ExSurv}
Figure 4 shows the process of implementing a sentiment analysis system, which are described below each one \cite{3}.\\

 \begin{figure*}
    \centering
    \includegraphics[width=17cm]{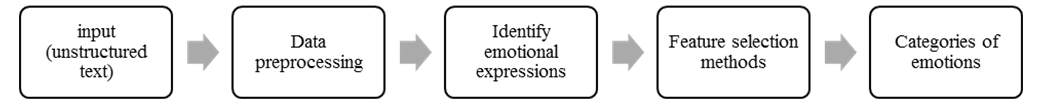}
    \caption{Implementation process of sentiment analysis system}
    \label{fig:life}
\end{figure*} 
\subsection{Input (unstructured text) }
The input data or dataset in sentiment analysis systems can be collected from social networks, news sites and forums using crawlers developed in programming languages such as Python; researchers should note that data or the extracted texts can analyze emotions. For example, users' recorded feedback about election candidates can be a good feed and product as input for sentiment analysis systems \cite{g1}. Also, in the input phase, researchers can choose their input data from datasets developed by other researchers, these types of datasets are available for free on sites such as Kaggle.
\subsection{Data preprocessing}
The data collected in the previous stage generally have non-printing characters, numbers, email, etc., which do not have any emotional meaning. Therefore, this type of data can be removed from the input text so that the system can perform the sentiment analysis process with better speed and accuracy \cite{a7}. Using natural language processing libraries such as NLTK, Spacy and TextBlob, researchers can easily perform part of the text preprocessing processes. Among the operations that are used in the pre-processing of text or unstructured text data, are:\\
• Stop word removal\\
• stemming and lemmatization\\
• Part Of Speech tagging\\
• Normalization\\
\subsection{Identify emotional expressions}
This stage is responsible for identifying emotional words. Therefore, the output of this step is a list of words (with emotional load) in the pre-processed text.

\subsection{Feature selection methods}
In this step, it is checked whether the sentences in the text have implicit and objective features or not. In section 3, various types of feature selection methods are discussed. In the following, with two different examples, implicit and objective features are explained.\\
Example 1: The sound quality of this phone is very good. It has an objective feature: "sound quality" and the emotional word "better" (positive feeling) \cite{g2}.\\
Example 2: This mobile is suitable for my pocket. It has an implicit feature: "mobile size" and the emotional word "appropriate" (positive feeling).\\
The details of this step are explained below.

\subsection{Categories of emotions}
In this step, machine learning algorithms (supervised and unsupervised) are used to categorize the sentiments of the records in the dataset according to the type of dataset and the values of the tag attributes \cite{b6}. Section 3 discusses various classification methods.
\section{Methods of feature selection and emotion classification}
Sentiment analysis is generally viewed as a classification problem, so text extraction and selection of text features is an important step to perform the classification process. In the following, some features that can be extracted from the text are described \cite{4}.\\
1. The presence of the word\\
Since the input data in sentiment analysis systems is very large and massive, the best thing is to extract appropriate features such as unique words (word tokenize), sets of words that are repeated with each other (n-gram), along with its frequency. specifying the presence of a word in the document in binary form (0 if not - 1 if present) can be suitable features for sentiment analysis of input documents \cite{5}.\\
2. Word frequency\\
Word frequency means the number of occurrences of a word or words in a document. Its mathematical symbol is as follows.\\

Another extractable feature is tf-idf, which indicates the number of times a word is repeated in the entire document or dataset.\\
3. Part of Speech Tagging\\
This type of feature recognizes the role of each word in the sentence, some of these roles are (noun, verb, adjective, adverb, etc.) The main idea of this feature is to recognize the feelings of a sentence according to the role Words are like adjectives. For example, it was a very good night. The word "good" as an adjective can give the sentence a positive emotion analysis tag.\\
4. The opinion of words\\
The purpose of extracting the meaning of words is the same attributes in the text. Words like (good, bad, ugly, etc.)\\

\subsection{Feature selection methods}
Figure 5 shows the types of feature selection methods in sentiment analysis systems\cite{6}.\\
\begin{figure}
    \centering
    \includegraphics[width=9cm,height=7cm]{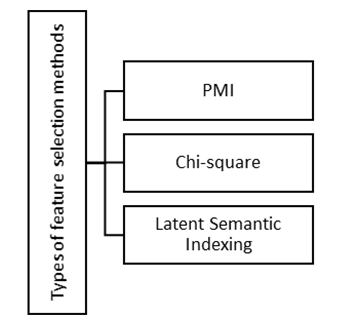}
    \caption{Types of feature selection methods}
    \label{fig:life}
\end{figure} 
1. PMI (Partial Mutual Information)\\
The PMI feature selection method is an unsupervised learning method derived by the work of \cite{7,b4} which is used to model the mutual information between features and classes. PMI algorithm is one of the most important concepts in natural language processing, which measures the probability of occurrence of two words with each other compared to random occurrence \cite{8}.\\
2. Chi-square method\\
If n is the total number of documents in the dataset or corpus. Pi(w) is called the conditional probability of the class or label i for documents that contain the word w. The chi-square method is better than the PMI method, because it gives more importance to the occurrences of words than each other \cite{9}.\\
3. Latent Semantic Indexing method\\
The main focus of feature selection methods is on reducing the dimensionality of the dataset. Most feature transformation methods focus on creating a smaller dataset of features from the original dataset. The LSI method uses a technique called singular value to find relational patterns between words and concepts in an unstructured text \cite{10}. LSI is one of the most famous feature transformation methods. LSI seeks to transform the text space into a linear combination of the features of the main words of the text. The main disadvantage of the LSI method is the lack of recognition and application of label features in the primary data set, because it was an unsupervised learning method \cite{11}.
\subsection{Emotional category}
Emotional classification refers to classifying documents based on two subjective and objective approaches, and in each of these approaches, word-based methods and machine learning methods can also be used.
\subsection{Machine learning methods}
Machine learning approaches are divided into two categories: supervised and unsupervised learning. Supervised learning methods use a dataset with a label feature. While in unsupervised learning methods, it is used when the label feature is not in the data set or is very difficult to obtain \cite{3}.
\subsubsection{Types of supervised learning algorithms}
The premise of using supervised learning algorithms is the existence of a data set with a label feature. Some of these algorithms are explained below.
1. Naïve Bayes\\
Simple Bayes algorithm uses Bayes theory to predict a label. This algorithm checks the probability of assigning a class to a set of words based on bow (p(label | feature)). The default of the Bayes algorithm refers to the independence of the features from each other, and the most important disadvantage is its use for small data sets \cite{3}.\\
2. Bayesian network algorithm\\
The default Bayesian network algorithm is based on two modes, either all features are interdependent or completely independent of each other. This network is like a graph whose nodes show the same features and edges show their dependencies. Bayesian network is usually used very little for sentiment analysis \cite{12}.\\
3. Maximum entropy algorithm\\
This algorithm is also called the conditional exponential classification algorithm, which uses a cryptographic method to convert the set of features with labels into vectors, and then each feature is given a weight to obtain its associated label \cite{a6}.\\
4. Linear bundles\\
In this section, the review and introduction of linear classification algorithms that classify labels or classes through the best separation line have been discussed.\\
5. Support vector machine algorithm\\
The svm algorithm is divided into two categories, linear and non-linear, and is one of the supervised learning algorithms. The svm algorithm is mostly used on textual data with a low number of features to perform the best separation between the data using the hyperplane \cite{13,14}.\\
6. Neural network algorithm\\
An artificial neural network consists of three layers: input, output and processing. Each layer contains a group of nerve cells (neurons) that are generally connected to all neurons in other layers, unless the user restricts the communication between neurons. Still, the neurons of each layer have no connection with other neurons of the same layer. Therefore, the sentiment analysis process can be done using neural network algorithms \cite{b1}.\\
7. Decision tree algorithms\\
These algorithms use a hierarchical tree to display training data, where each feature is in the role of a node and its values are in the role of edges of that node \cite{15}. This hierarchical structure continues until it reaches ninety pages. Each ninety leaves is a label \cite{16}. Various criteria are used to create a hierarchical structure in decision tree algorithms. It is a suitable criterion to perform the best differentiation through the selection of dataset features (in the role of tree nodes) \cite{17,b2}.\\
8. Rule-based algorithms\\
In these algorithms, a set of rules are used to model the data space. Two important criteria in this type of algorithms are reliability and trust factor criteria, which create rules based on the number of occurrences of items in a dataset. Then, the words that have the specified minimum threshold in two confidence and confidence coefficients are selected for classification \cite{3}.
\subsubsection{Unsupervised learning algorithms}
Another method of machine learning is the unsupervised learning method. When a training dataset with correct labels cannot be accessed, supervised learning methods are used to classify documents. In this method, the algorithm starts learning by using the similarities and patterns between the data and finally categorizes the documents. Word-based approaches are among these types of algorithms that are explained below \cite{12}.
\subsubsection{Word-based approaches}
Most of the input data used in the sentiment analysis process consists of a number of opinion or sentiment words. For example, most of the comments recorded about a political news or the sale of a product can have these words, by examining them, a positive or negative label can be assigned to a document. In the following, some of them are examined \cite{3}.\\
1. Dictionary-based approaches\\
In this approach, an emotional word dictionary is used, which can be assigned a positive or negative label in the analysis of emotions by checking the words of a sentence with the words in the dictionary. This dictionary can grow gradually over the course of the classification, like a wordnet that has a collection of synonyms and antonyms for a word. But the problem that exists in this type of approach is that the dictionary does not fully understand the meanings of the words.\\
For example: I got a lion and took it home. In this example, it is not possible to understand which milk is meant by milk (animal, tap, drinking milk).\\
2. Corpus based approaches\\
This approach uses a large corpus of words to overcome the problem of dictionary-based approaches regarding syntactic patterns of word usage. On Dictionary-based Kurds were not able to extract the meaning of words according to the field of its application. These approaches require a large collection of words used in a field, so they have no advantage over dictionary-based methods \cite{17}.
\subsubsection{Semantic approaches}
1. Natural and word-based language processing techniques\\
In this method, in addition to word-based methods, natural language processing techniques such as word component tag extraction are also used to extract the syntactic structure of words in a sentence in addition to the semantic relationship \cite{18,19}.\\
2. Discourse information\\
Discourse information has a special application and importance in sentiment analysis systems. This information can be between sentences or words in the sentence. \cite{19} suggests five types of relationships between words, which are respectively contrast, correlation, support, conclusion and continuation, which are attached with emotional information and annotation. In the work of \cite{20}, he used discourse information to remove ambiguity in polarity and filter the polarity term to identify the relationship.\\
3. Word-based approaches\\
In this method, a list of words or an emotional dictionary is used, which has a collection of different emotional words, then by checking the words in the text and comparing it with the words in the dictionary, it is decided on the type of "positive" emotions in the text. And "negative" is taken.
\section{Applications of sentiment analysis}
Some of the applications of sentiment analysis systems include online advertising, analyzing the attitude of blog writers, analyzing the attitude of people towards a product, etc. \cite{21}. In the following, more applications of sentiment analysis systems are explained.\\
1. Application of sentiment analysis in recommender systems\\
People mostly register their opinions about a product on the stores website. Therefore, it is possible to develop a system that, according to the sentiment analysis of user comments recorded on store websites, recommends higher quality and better rated products to new website users.\\
2. Application of sentiment analysis in advertising\\
Many websites dedicate a part of their site to online advertising. An online marketer can choose a website that has better content for advertising their products by analyzing the content of advertising websites. Because it is possible, by choosing a website with inappropriate content, it will be penalized by search engines and have limited income.\\
3. Application of sentiment analysis in business intelligence\\
In business intelligence, the goal is to increase the profitability of a business and better communication with customers through data analysis. Text comments registered on a website is a set of unstructured text data that a business intelligence expert can provide a better way to register positive comments and, as a result, better profit from product sales by analyzing the sentiments of user comments.\\
4. Application of sentiment analysis in trend forecasting\\
One of the most important applications of sentiment analysis systems is their use in predicting various trends.\\
For example, a trader can make a more informed decision about buying or selling a digital currency by analyzing the sentiments of opinions recorded about it, or a group of mountain climbers can predict the weather by examining the opinions recorded by other mountain climbers. A mountainous area like the Everest mountain range.\\
5. The application of sentiment analysis in the field of politics\\
Another application of sentiment analysis systems is its application in the field of politics, such as predicting the election process, checking the level of user satisfaction with determined policies, etc.\\
For example, a president can have a report on the level of people's satisfaction with his ministers by analyzing the sentiments of users' opinions.\\
6. Application of sentiment analysis in smart home\\
A smart home can make a better decision based on the feeling of recorded or expressed opinions of the people of the house.\\
For example, if a person in a smart home says "how hot is the air in the house", the smart home system can activate the cooling system by analyzing the user's opinion.\\
\section{Challenges in sentiment analysis}
When developing a sentiment analysis system, the researcher should pay attention to the challenges described below \cite{a1,22,a3}.\\
1. The challenge of identifying the subjective part of a text document\\
When examining words in sentiment analysis, it is possible that some words are extracted as a mental item in one text, while they are used as an objective item in another document.\\
Example: The author's language is very crude.
The system may have the extraction of crude oil, including the above, so mentally, the extraction has not been done correctly.\\
2. The challenge of communicating feelings with specific words\\
Sometimes it is very easy to get the feeling of a sentence, but it can be difficult to get its emotional source \cite{23,w4,a5}.\\
For example, the sentence: I was very happy to see them. This sentence has a positive feeling, but the source of that "they" feeling is not clear.\\
3. The challenge of detecting sarcasm\\
Sometimes, the sentiment of a sentence may be incorrectly extracted and the sentiment analysis system does not notice the sarcasm or sarcasm of the words in it.\\
For example: He has gone from extreme ugliness to indescribable beauty. In this example, the sentiment analysis system may incorrectly assign the tag "positive" to the sentence.\\
4. Neutral phrase challenge\\
Sometimes the minority of the text can affect the emotion extraction of a text \cite{23}.\\
5. The challenge of extracting negative emotions indirectly\\
Sometimes, the presence of a series of words (such as no, refrain, etc \cite{w5,a4}.) in the text creates a wrong and indirect feeling.\\
For example: from eating while driving please The sentiment analysis system may place the sentence in the negative category due to the presence of the word refrain.\\
6. The challenge of order dependence\\
Another challenge of sentiment analysis systems is not giving importance to the relationship and placement of words \cite{12}.\\
For example: A car is better than a motor. For this sentence, the sentiment analysis system may identify the engine better than the car.\\
7. Entity identification challenge\\
The use of the BOW method in sentiment analysis systems may not have any recognition of the entities in the text.\\
For example: I like a Samsung phone, but I use an iPhone. The sentiment analysis system cannot recognize that the two words Samsung and iPhone are two entities in this sentence.\\
8. The challenge of using problematic text\\
If the input text has wrong words in terms of grammar and spelling, the sentiment analysis system may not be able to recognize the emotional tag of the text.\\
For example: I like the car with yellow color, because I like it, it is wrongly typed, the sentiment analysis system will not have a correct understanding of the feeling of this sentence.\\
9. The challenge of languages\\
Algorithms and tools for the development of sentiment analysis systems are mostly developed for English and Chinese languages. Therefore, analyzing the emotions of Persian texts is a big challenge.\\
10. The challenge of detecting spam comments\\
In some cases, people with the intention of ruining a company's product, register negative comments about its products \cite{w2}. The sentiment analysis system is not able to recognize spam texts and may mistakenly label a product as negative \cite{b3}.\\
11. The challenge of identifying the expert\\
The text in a document is not always a quote from a specific person, sometimes the author of a text may quote a sentence from another person, in which case sentiment analysis systems cannot identify the original source and author of that quote recognize \cite{w3}.\\
\bibliographystyle{IEEEtran}
\bibliography{References}

\begin{thebibliography}{10}
\providecommand{\url}[1]{#1}
\csname url@samestyle\endcsname
\providecommand{\newblock}{\relax}
\providecommand{\bibinfo}[2]{#2}
\providecommand{\BIBentrySTDinterwordspacing}{\spaceskip=0pt\relax}
\providecommand{\BIBentryALTinterwordstretchfactor}{4}
\providecommand{\BIBentryALTinterwordspacing}{\spaceskip=\fontdimen2\font plus
\BIBentryALTinterwordstretchfactor\fontdimen3\font minus
  \fontdimen4\font\relax}
\providecommand{\BIBforeignlanguage}[2]{{%
\expandafter\ifx\csname l@#1\endcsname\relax
\typeout{** WARNING: IEEEtran.bst: No hyphenation pattern has been}%
\typeout{** loaded for the language `#1'. Using the pattern for}%
\typeout{** the default language instead.}%
\else
\language=\csname l@#1\endcsname
\fi
#2}}
\providecommand{\BIBdecl}{\relax}
\BIBdecl

\bibitem{w1}
M.~Kitsa, ``The use of social networks by british media the telegraph and bbc
  news,'' \emph{State and Regions. Series: Social Communications}, no. 1 (49),
  pp. 80--86, 2022.

\bibitem{1}
B.~Liu, ``Sentiment analysis and opinion mining. synthesis lectures on human
  language technologies, 5 (1), 1-167,'' 2012.

\bibitem{2}
T.~Wilson, J.~Wiebe, and P.~Hoffmann, ``Recognizing contextual polarity in
  phrase-level sentiment analysis,'' in \emph{Proceedings of human language
  technology conference and conference on empirical methods in natural language
  processing}, 2005, pp. 347--354.

\bibitem{a1}
A.~Yazdinejad, R.~M. Parizi, G.~Srivastava, and A.~Dehghantanha, ``Making sense
  of blockchain for ai deepfakes technology,'' in \emph{2020 IEEE Globecom
  Workshops (GC Wkshps}.\hskip 1em plus 0.5em minus 0.4em\relax IEEE, 2020, pp.
  1--6.

\bibitem{3}
W.~Medhat, A.~Hassan, and H.~Korashy, ``Sentiment analysis algorithms and
  applications: A survey,'' \emph{Ain Shams engineering journal}, vol.~5,
  no.~4, pp. 1093--1113, 2014.

\bibitem{g1}
K.~Viswanathan and A.~Yazdinejad, ``Security considerations for virtual reality
  systems,'' \emph{arXiv preprint arXiv:2201.02563}, 2022.

\bibitem{a7}
E.~Rabieinejad, A.~Yazdinejad, and R.~M. Parizi, ``A deep learning model for
  threat hunting in ethereum blockchain,'' in \emph{2021 IEEE 20th
  International Conference on Trust, Security and Privacy in Computing and
  Communications (TrustCom)}.\hskip 1em plus 0.5em minus 0.4em\relax IEEE,
  2021, pp. 1185--1190.

\bibitem{g2}
S.~Nakhodchi, B.~Zolfaghari, A.~Yazdinejad, and A.~Dehghantanha, ``Steeleye: An
  application-layer attack detection and attribution model in industrial
  control systems using semi-deep learning,'' in \emph{2021 18th International
  Conference on Privacy, Security and Trust (PST)}.\hskip 1em plus 0.5em minus
  0.4em\relax IEEE, 2021, pp. 1--8.

\bibitem{b6}
D.~Sheridan, J.~Harris, F.~Wear, J.~Cowell~Jr, E.~Wong, and A.~Yazdinejad,
  ``Web3 challenges and opportunities for the market,'' \emph{arXiv preprint
  arXiv:2209.02446}, 2022.

\bibitem{4}
M.~Vijyalaxmi, S.~Chopra, S.~Oswal, M.~D. Chaturvedi \emph{et~al.}, ``The how,
  when and why of sentiment analysis,'' \emph{International Journal of Computer
  Technology and Applications}, vol.~4, no.~4, p. 660, 2013.

\bibitem{5}
Y.~Mejova and P.~Srinivasan, ``Exploring feature definition and selection for
  sentiment classifiers,'' in \emph{Proceedings of the International AAAI
  Conference on Web and Social Media}, vol.~5, no.~1, 2011, pp. 546--549.

\bibitem{6}
C.~Whitelaw, N.~Garg, and S.~Argamon, ``Using appraisal groups for sentiment
  analysis,'' in \emph{Proceedings of the 14th ACM international conference on
  Information and knowledge management}, 2005, pp. 625--631.

\bibitem{7}
T.~M. Cover, \emph{Elements of information theory}.\hskip 1em plus 0.5em minus
  0.4em\relax John Wiley \& Sons, 1999.

\bibitem{b4}
A.~Yazdinejad, R.~M. Parizi, A.~Dehghantanha, H.~Karimipour, G.~Srivastava, and
  M.~Aledhari, ``Enabling drones in the internet of things with decentralized
  blockchain-based security,'' \emph{IEEE Internet of Things Journal}, vol.~8,
  no.~8, pp. 6406--6415, 2020.

\bibitem{8}
N.~F. Da~Silva, E.~R. Hruschka, and E.~R. Hruschka~Jr, ``Tweet sentiment
  analysis with classifier ensembles,'' \emph{Decision support systems},
  vol.~66, pp. 170--179, 2014.

\bibitem{9}
C.~C. Aggarwal and C.~C. Aggarwal, \emph{Mining text data}.\hskip 1em plus
  0.5em minus 0.4em\relax Springer, 2015.

\bibitem{10}
S.~B. Bhonde and J.~R. Prasad, ``Sentiment analysis-methods, applications \&
  challenges,'' \emph{International Journal of Electronics Communication and
  Computer Engineering}, vol.~6, no.~6, p. 634, 2015.

\bibitem{11}
X.~Zheng, Z.~Lin, X.~Wang, K.-J. Lin, and M.~Song, ``Incorporating appraisal
  expression patterns into topic modeling for aspect and sentiment word
  identification,'' \emph{Knowledge-Based Systems}, vol.~61, pp. 29--47, 2014.

\bibitem{12}
D.~Maynard and A.~Funk, ``Automatic detection of political opinions in
  tweets,'' in \emph{The Semantic Web: ESWC 2011 Workshops: ESWC 2011
  Workshops, Heraklion, Greece, May 29-30, 2011, Revised Selected Papers
  8}.\hskip 1em plus 0.5em minus 0.4em\relax Springer, 2012, pp. 88--99.

\bibitem{a6}
A.~Yazdinejad, R.~M. Parizi, A.~Dehghantanha, and K.-K.~R. Choo,
  ``P4-to-blockchain: A secure blockchain-enabled packet parser for software
  defined networking,'' \emph{Computers \& Security}, vol.~88, p. 101629, 2020.

\bibitem{13}
C.~Cortes and V.~Vapnik, ``Support-vector networks,'' \emph{Machine learning},
  vol.~20, pp. 273--297, 1995.

\bibitem{14}
V.~Vapnik, \emph{The nature of statistical learning theory}.\hskip 1em plus
  0.5em minus 0.4em\relax Springer science \& business media, 1999.

\bibitem{b1}
A.~Yazdinejad, B.~Zolfaghari, A.~Dehghantanha, H.~Karimipour, G.~Srivastava,
  and R.~M. Parizi, ``Accurate threat hunting in industrial internet of things
  edge devices,'' \emph{Digital Communications and Networks}, 2022.

\bibitem{15}
J.~Quinlan, ``Introduction of decision tree: Machine learn,'' 1986.

\bibitem{16}
S.~Chakrabarti, S.~Roy, and M.~V. Soundalgekar, ``Fast and accurate text
  classification via multiple linear discriminant projections,'' \emph{The VLDB
  journal}, vol.~12, pp. 170--185, 2003.

\bibitem{17}
D.~D. Lewis and M.~Ringuette, ``A comparison of two learning algorithms for
  text categorization,'' in \emph{Third annual symposium on document analysis
  and information retrieval}, vol.~33, 1994, pp. 81--93.

\bibitem{b2}
A.~Yazdinejad, M.~Kazemi, R.~M. Parizi, A.~Dehghantanha, and H.~Karimipour,
  ``An ensemble deep learning model for cyber threat hunting in industrial
  internet of things,'' \emph{Digital Communications and Networks}, 2022.

\bibitem{18}
S.-M. Kim and E.~Hovy, ``Determining the sentiment of opinions,'' in
  \emph{COLING 2004: Proceedings of the 20th International Conference on
  Computational Linguistics}, 2004, pp. 1367--1373.

\bibitem{19}
N.~Asher, F.~Benamara, and Y.~Y. Mathieu, ``Distilling opinion in discourse: A
  preliminary study,'' in \emph{22nd International Conference on Computational
  Linguistics (COLING 2008)}.\hskip 1em plus 0.5em minus 0.4em\relax ACL:
  Association for Computational Linguistics, 2008, pp. 7--10.

\bibitem{20}
A.~Moreo, M.~Romero, J.~Castro, and J.~M. Zurita, ``Lexicon-based
  comments-oriented news sentiment analyzer system,'' \emph{Expert Systems with
  Applications}, vol.~39, no.~10, pp. 9166--9180, 2012.

\bibitem{21}
D.~Osimo and F.~Mureddu, ``Research challenge on opinion mining and sentiment
  analysis,'' \emph{Universite de Paris-Sud, Laboratoire LIMSI-CNRS,
  B{\^a}timent}, vol. 508, 2012.

\bibitem{22}
F.~Xianghua, L.~Guo, G.~Yanyan, and W.~Zhiqiang, ``Multi-aspect sentiment
  analysis for chinese online social reviews based on topic modeling and hownet
  lexicon,'' \emph{Knowledge-Based Systems}, vol.~37, pp. 186--195, 2013.

\bibitem{a3}
A.~Yazdinejad, A.~Dehghantanha, R.~M. Parizi, and G.~Epiphaniou, ``An optimized
  fuzzy deep learning model for data classification based on nsga-ii,''
  \emph{Neurocomputing}, vol. 522, pp. 116--128, 2023.

\bibitem{23}
G.~Jaganadh, ``Opinion mining and sentiment analysis,'' \emph{CSI
  Communications}, 2012.

\bibitem{w4}
H.~Li, X.~Bruce, G.~Li, and H.~Gao, ``Restaurant survival prediction using
  customer-generated content: An aspect-based sentiment analysis of online
  reviews,'' \emph{Tourism Management}, vol.~96, p. 104707, 2023.

\bibitem{a5}
A.~Yazdinejad, A.~Dehghantanha, R.~M. Parizi, M.~Hammoudeh, H.~Karimipour, and
  G.~Srivastava, ``Block hunter: Federated learning for cyber threat hunting in
  blockchain-based iiot networks,'' \emph{IEEE Transactions on Industrial
  Informatics}, vol.~18, no.~11, pp. 8356--8366, 2022.

\bibitem{w5}
L.~Shang, H.~Xi, J.~Hua, H.~Tang, and J.~Zhou, ``A lexicon enhanced
  collaborative network for targeted financial sentiment analysis,''
  \emph{Information Processing \& Management}, vol.~60, no.~2, p. 103187, 2023.

\bibitem{a4}
A.~Yazdinejad, R.~M. Parizi, A.~Dehghantanha, Q.~Zhang, and K.-K.~R. Choo, ``An
  energy-efficient sdn controller architecture for iot networks with
  blockchain-based security,'' \emph{IEEE Transactions on Services Computing},
  vol.~13, no.~4, pp. 625--638, 2020.

\bibitem{w2}
C.~Van~Dinh, S.~T. Luu, and A.~G.-T. Nguyen, ``Detecting spam reviews on
  vietnamese e-commerce websites,'' in \emph{Intelligent Information and
  Database Systems: 14th Asian Conference, ACIIDS 2022, Ho Chi Minh City,
  Vietnam, November 28--30, 2022, Proceedings, Part I}.\hskip 1em plus 0.5em
  minus 0.4em\relax Springer, 2022, pp. 595--607.

\bibitem{b3}
A.~Yazdinejad, R.~M. Parizi, A.~Dehghantanha, and K.-K.~R. Choo,
  ``Blockchain-enabled authentication handover with efficient privacy
  protection in sdn-based 5g networks,'' \emph{IEEE Transactions on Network
  Science and Engineering}, vol.~8, no.~2, pp. 1120--1132, 2019.

\bibitem{w3}
M.~Aliarab and K.~Fouladi, ``A survey on review spam detection methods using
  deep learning approach,'' \emph{International Journal of Web Research},
  vol.~5, no.~1, pp. 19--24, 2022.

\end{thebibliography}
\end{document}